# Characterization of Dynamic Bayesian Network
## The Dynamic Bayesian Network as temporal network

Nabil Ghanmi
National School of Engineer of Sousse
Sousse - Tunisia
Nabil.ghnamy@gmail.com

Mohamed Ali Mahjoub
Preparatory Institute of Engineer of Monastir
Monastir - Tunisia
medali.mahjoub@ipeim.rnu.tn

Najoua Essoukri Ben Amara
National School of Engineer of Sousse
Sousse - Tunisia
Najoua.benamara@eniso.rnu.tn

*Abstract*— In this report, we will be interested at Dynamic Bayesian Network (DBNs) as a model that tries to incorporate temporal dimension with uncertainty. We start with basics of DBN where we especially focus in Inference and Learning concepts and algorithms. Then we will present different levels and methods of creating DBNs as well as approaches of incorporating temporal dimension in static Bayesian network.

## I. INTRODUCTION

The majority of events encountered in everyday life are not well described based on their occurrence at a particular point in time but rather they are described by a set of observations that can produce a comprehensive final event. Thus, time is an important dimension to take into account in reasoning and in the field of artificial intelligence in general. To add the time dimension in Bayesian networks, different approaches have been proposed. The common names used to describe this new dimension are "temporal" and "dynamic ".

## II. BASICS

### A. Definition

Bayesian networks represent a set of variables in the form of nodes on a directed acyclic graph. It maps the conditional independencies of these variables. They bring us four advantages as a data modeling tool [16,17,18]

A dynamic Bayesian network can be defined as a repetition of conventional networks in which we add a causal one time step to another. Each Network contains a number of random variables representing observations and hidden states of the process.

We consider a dynamic Bayesian network composed of a sequence of T hidden state variables (a hidden state of a DBN is represented by a set of hidden state variables) and a sequence of T observable variables where T is time limit of the studied process.

In order that the specification of this network is complete, we need to define the following parameters:

- The transition probability between states $P(x_t/x_{t-1})$
- The conditional probability of hidden states knowing observation $P(y_t/x_t)$
- The probability of the initial state $P(x_1)$

The first two parameters must be determined for each time $t = 1, \ldots, T$. These parameters can be invariant or not over time.

### B. Inference

The general problem of inference for DBNs is to calculate $P(X_{t0}^i/Y_{t1:t2})$ where $X_{t0}^i$ is the $i^{th}$ hidden variable at time $t0$ and $Y_{t1:t2}$ represents all observations between times $t1$ and $t2$.

There are several interesting cases of inference, they are illustrated below. The arrow indicates $t0$: that we try to estimate. Shaded regions correspond to observations between $t1$ and $t2$

**Filtering**: this is to estimate the belief state at time $t0$ knowing all the observations until this moment:

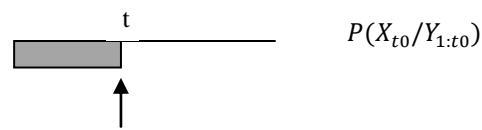

$P(X_{t0}/Y_{1:t0})$

**Decoding (Viterbi)**: decoding problem is to determine the most likely sequence of hidden states knowing the observations up to time $t0$:

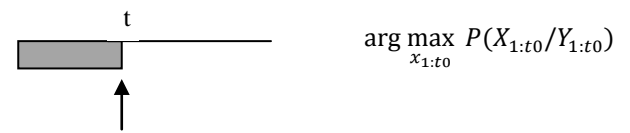

$\arg\max_{x_{1:t0}} P(X_{1:t0}/Y_{1:t0})$

**Prediction:** This is to estimate a future observation or state knowing the observations up to the current time t0

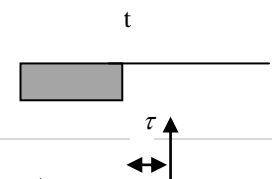

$$P(X_{t0+\tau}/Y_{1:t0})$$

**Smoothing (offline)**: is to estimate a past state knowing the observations up to the current time T

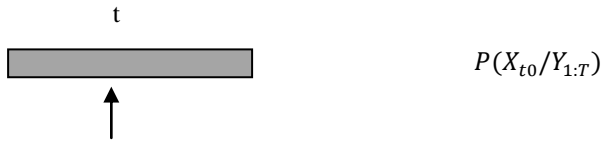

$$P(X_{t0}/Y_{1:T})$$

There are several algorithms for inference in Dynamic Bayesian Networks. We can classify these algorithms according to their accuracy, in two broad classes:

*1) Exact Inference:*

*a) Forward-Backward Algorithm:*

The algorithm proceeds in two steps:

1) Forward step: forward propagation of probabilities

2) Backward step: backward propagation of probabilities

• **Forward Algorithm:**

We consider a dynamic Bayesian network B. We wish to calculate the probability $P(Y_{1:T})$ of occurrence of the sequence of observation $Y_{1:T}$. This probability is:

$$P(Y_{1:T}) = \sum_{\substack{possible \\ paths}} \left[ \left( \prod_{t=1}^{T-1} P(x_{t+1}/x_t) P(y_{t+1}/x_t) \right) P(x_{T+1}/x_T) \right] \quad (1)$$

Applying directly this formula, the computation time is O(TNT). For this, we consider the forward variable defined by:

$$\alpha_t(x_t) = P(Y_{1:t}, x_t) \quad (2)$$

which expresses the probability of observing the sequence $Y_{1:t}$ while lying in state $x_t$. This variable can be computed inductively:

Initialisation :

$$\alpha_1(x_1) = P(x_1)$$

Induction :

$$\alpha_{t+1}(x_{t+1}) = P(y_{t+1}/x_{t+1}) \left( \sum_{x_t} P(x_{t+1}/x_t) \alpha_t(x_t) \right) \quad (3)$$

Thus, we can calculate $\alpha_T(x_T) = P(Y_{1:T}, x_T)$, this naturally leads us to:

$$P(Y_{1:T}) = \frac{\alpha_T(x_T)}{\sum_{x_t} \alpha_t(x_t)} \quad (4)$$

• **Backward Algorithm:**

It is also possible to perform the calculation in reverse, using the backward algorithm.

For this, we define the backward variable as follows:

$$\beta_t(x_t) = P(Y_{t+1:T}/x_t) \quad (5)$$

This variable expresses the conditional probability of observation from time t +1 until the last observation time T, given the values of the hidden states at time t. Its calculation follows the following procedure:

Initialisation :

$$\beta_T(x_T) = 1$$

Induction :

$$\beta_{t-1}(x_{t-1}) = \sum_{x_t} P(x_t/x_{t-1}) . P(y_t/x_t) . \beta_t(x_t) \quad (6)$$

Thus, we can calculate the expected probability:

$$P(Y_{1:T}) = \sum_{x_1} \beta_1(x_1) . P(y_1/x_1) . \quad (7)$$

The complexity of this algorithm is, as the forward algorithm in O(TN$^2$).

From these two factors (forward and backward) propagation probabilities, we can explore other terms that are useful for inference and learning of Dynamic Bayesian networks:

- **Smoothing**: this is to calculate $P(x_t/Y_{1:T})$ where t < T. From equations (4) and (6), we can determine the following equation:

$$\gamma_t = P(x_t/Y_{1:T}) = \frac{\alpha_t(x_t)\beta_t(x_t)}{\sum_{x_t} \alpha_t(x_t)\beta_t(x_t)} \quad (8)$$

$\gamma_t$ is called smoothing operator. We can also derive higher order smoothing equations. For example, a smoothing of the first order is defined as follows:

$$\xi_{k,k-1}(x_t, x_{t-1}) = P(x_t, x_{t-1}/Y_{1:T})$$
$$= \frac{\alpha_{t-1}(x_{t-1})P(x_t, x_{t-1})P(y_t/x_t)\beta_t(x_t)}{\sum_{x_t} \alpha_t(x_t)\beta_t(x_t)} \quad (9)$$

These terms may be used to easily calculate the probabilities of hidden states from the neighboring nodes.

- **Prediction**: this is to calculate $P(x_{t+1}/Y_{1:t})$ et $P(y_{t+1}/Y_{1:T})$. We can easily determine:

$$P(x_{t+1}/Y_{1:t}) = \frac{\sum_{x_t} P(x_{t+1}/x_t) \alpha_t(x_t)}{\sum_{x_t} \alpha_t(x_t)} \quad (10)$$

Similarly, one can determine:

$$P(y_{t+1}/Y_{1:t}) = \frac{\sum_{x_{t+1}} \alpha_{t+1}(x_{t+1})}{\sum_{x_t} \alpha_t(x_t)} \qquad (11)$$

- **Decoding** : is to determine the sequence of hidden states $\hat{X}_{1:T}$ such as :

$$\hat{X}_{1:T} = \arg\max_{X_{1:T}} P(X_{1:T}/Y_{1:T}) \qquad (12)$$

This task can be solved using the dynamic programming algorithm of Viterbi. We can start with the following equation:

$$\delta_{t+1}(x_{t+1}) = \max_{X_{1:t}} P(X_{1:t+1}/Y_{1:t+1}) \qquad (13)$$

Considering the topology of the DBN, we can deduce:

$$\delta_{t+1}(x_{t+1}) = P(y_{t+1}/x_{t+1}) \max_{x_t}\left[ P(x_{t+1}/x_t) \max_{x_{1:t-1}} P(X_{1:t}/Y_{1:t}) \right]$$
$$= P(y_{t+1}/x_{t+1}) \max_{x_t}[P(x_{t+1}/x_t)\delta_t(x_t)] \qquad (14)$$

We can now easily deduce that:

$$\max_{x_{1:T-1}} P(X_{1:T-1}/Y_{1:T-1}) = \max_{x_{T-1}} \delta_{T-1}(x_{T-1}) \qquad (15)$$

To find $\hat{X}_{1:T}$, we must introduce the argument $x_t$ that maximizes $\delta_{t+1}(x_{t+1})$ as follows:

$$\psi_{t+1}(x_{t+1}) = \arg\max_{X_t}[P(x_{t+1}/x_t) + \delta_t(x_t)] \qquad (16)$$

And we have:

$$\hat{x}_t = \psi_{t+1}(\hat{x}_{t+1}) \qquad (17)$$

Note that if we want to use the Viterbi algorithm to decode the sequence of hidden states, we must have a complete observation $y_1, \dots, y_T$. If the number of observations is not sufficient, a less optimal solution known as the truncated Viterbi algorithm can be used.

*b) Junction Tree Algorithm:*

The Junction Tree Algorithm **[1]** is an algorithm similar to the Baum-Welch algorithm used in HMM. It involves transforming the original network into a new structure called junction tree and apply a type inference algorithm used for static Bayesian networks. This tree is obtained by following these steps:

- **Moralization**: connecting parents and eliminating directions.
- **Triangularization**: selectively adding arcs to the graph morale (not to have cycles of order 4 or more).
- **Junction Tree** : is obtained from the triangulated graph by connecting the cliques such that all cliques on the path between two cliques X and Y contain X ∩ Y

*2) Approximate inference:*

When the dimension of Bayesian networks increases, the computing time is increasingly important. When the conditional probability tables are derived from data (learning), these tables are not accurate. In this case it is not worth wasting time by making exact inference on probabilities not precise, hence the use of approximate inference methods. Among the approximate inference methods that often work well in practice, we give:

*a) Variational methods*

The simplest example is the approximation by the average (mean-field approximation) [2], which exploits the law of large numbers to approximate large sums of random variables by their average. The approximation by the average product of a lower probability. There are other, more stringent, resulting in a lower and upper.

*b) Monte Carlo*

The easiest Monte Carlo Method **[3]** is the Importance Sampling (IS) that produces a large number of samples x from $P(X)$ the unconditional distribution of hidden variables) then we give weight to samples based on their likelihood $P(Y/X)$ (where y is the observation). This forms the basis of Particulate Filter which is simply the Importance Sampling adapted to a dynamic Bayesian network.

*c) Loopy Belief propagation*

We apply the algorithm of Pearl **[4]** to the original graph even if it contains loops. In theory, one runs the risk of double counting certain words but it was shown that in some cases (for example, a single loop), events are counted twice and thus cancel out fairly between them to give the correct answer .

*C. Leaning:*

Learning is to estimate the probability tables and conditional distributions CPTs CPDS. This task is based on the EM algorithm (Expectation Maximization) algorithm or the GEM (General Expectation Maximization) for DBNs.

Let M be a Dynamic Bayesian network with parameter $\theta$, learning aims to determine $\hat{\theta}$ such the posterior probability of the observations is maximal, then either:

$$\hat{\theta} = \arg\max_\theta[P(Y,X/\theta] \qquad (18)$$

**EM Algorithm**: This algorithm includes:

- an evaluation step of expectation (E), which calculates the expectation of the likelihood taking into account the recent observed variables,
- a maximization step (M), where an estimated maximum likelihood parameters by maximizing the likelihood found in step E.

*D. Pruning*

This task is based on the possibility of change in time, of RBD's structure. This is usually omitted for its complexity. Pruning the network consists in perform one of the following operations:
- Delete one or more states of a given node
- Remove a connection between two nodes
- Remove one or more network nodes

This can be exact (lossless) or approximate

III. DIFFERENT LEVELS OF CREATING DBN

To describe a dynamic Bayesian network, we must specify its topology (the graph structure) as well as all the tables of conditional probability distribution. You can learn them both (the graph and distributions) from experimental data. However, it is more difficult to learn a structure to learn its parameters.

It is possible that some nodes are hidden during the experiments (values that we can't observe), or missing data. In this case, learning becomes more complicated settings. From these considerations, there are 4 possible cases of learning [5]:

TABLE I. METHODS OF DETERMINATION OF DBN STRUCTURE AND PARAMETERS

| Structure | Observability | Method |
|---|---|---|
| Known | Full | Simple statistics : MLE |
| Known | Partial | EM or Gradient Ascent Algorithm |
| Unknown | Full | Search through model space |
| Unknown | Partial | Structural EM |

*A. Known structure /Full observability*

The DBN's structure is known, it remains to estimate the parameters of the network using the method of maximum likelihood estimation. We look for parameters $\theta$ describing the model assumptions that maximize the likelihood of observations Y:

$$l(Y,\theta) = P(Y/\theta) \quad (19)$$

In general, it instead uses the log likelihood (log-likelihood)

$$L(Y,\theta) = \log ( P(Y/\theta)) \quad (20)$$

*B. Known structure /Partial observability*

When certain variables are not observable, the likelihood surface becomes multimodal and we must use iterative methods such as EM or gradient increasing to find local maxima of the function ML / MAP. The principle of the EM algorithm is to associate a problem with an incomplete data problem for which complete data for a simple solution exists for the maximum likelihood estimate. This procedure needs to use an inference algorithm to compute the parameters for each node. These algorithms are explained in section II.3

*C. Unknown structure / Full observability*

There are several techniques for learning DBN structure from observed data. These techniques help to create the network structure by adding or deleting edges between any two nodes or reversing the direction of an existing arc. These changes must be made in order to maintain and acyclic directed graph.

To accomplish the task of structural learning, we need [6]:
- an algorithm to find the different possible structures
- a metric for comparing the possible structures to each other

The structure learning algorithms can be classified into two broad categories.

- The first class of algorithms using heuristic search methods to construct the graph and evaluates it using scores (scoring methods). This procedure is repeated until the improvement between two consecutive models is not significant.
- The second class of algorithms to create the network structure by analyzing the independence relations between nodes. These independence relations are measured using several types of tests of conditional independence (eg mutual information between two nodes can be considered as a criterion for conditional independence)

According to Cheng et al. [7], when comparing the two types of algorithms, we can conclude that the first class of algorithms are faster than the second if the network is densely connected, but can't find the best solution for most models corresponding to real processes of the heuristic nature of these algorithms. The second class of algorithms can produce, under some assumptions, an optimal or near optimal solution especially when the data are not numerous.

### D. Unknown Structure /Partial Observability

The EM algorithm is developed to make learning network settings, so it must be adjusted to perform structural learning from incomplete data. The structural EM (SEM) is one of the most popular techniques that are developed for this purpose. SEM has the same E-step EM algorithm for completing the data using observations and the current structure of the network. The M-step involves two parts: In the first, it recalculates as already explained, the maximum likelihood to determine the parameters. In the second part, it uses these parameters to evaluate any other candidate structure similar to the current structure.

## IV. DIFFERENT APPROACHES FOR INCORPORATING TIME IN BAYESIAN NETWORK

Dynamic Bayesian Networks (DBN) are an extension of Bayesian networks that represent the temporal or spatial evolution of random variables. There are several models for incorporating time into network representation. These models can be classified into three broad categories:

- Models that use static BNs and formal grammars to represent the temporal dimension (temporal probabilistic networks (TPNs)
- Models that use a mixture of several probabilistic frameworks
- Models that use temporal nodes in the static BNs to represent temporal dependencies

The first two models are developed for specific objectives and have a very limited use. We will therefore focus on the third model.

### A. Probabilistic Temporal Networks (PTN)

#### 1) Definition

A probabilistic temporal network (PTN) is defined as a model, representing the time information while fully embracing probabilistic semantics. In a PTN, the nodes of the graph are the temporal aggregates and the arcs are causal and / or temporal relations

This type of network uses grammatical rules to express temporal dependencies in the structure of Bayesian networks: The conservation of the structure of static Bayesian networks allow reuse of powerful techniques for inference of BNs this specific type of networks. Grammar introduce temporal relations between events

#### 2) Temporal Reasoning

In PTN, temporal reasoning is based on interval algebra **[8]** which was introduced by James F. Allen in 1983. This is a calculation that defines the possible relationships between time intervals and provides a table of composition that can be used as a basis for reasoning on descriptions of temporal events.

The 13 following basic relations capture possible relationships between two intervals are illustrated in the following table:

TABLE II. ALLEN'S INTERVAL ALGEBRA

| Relation | Illustration | Interprétation |
|---|---|---|
| X < Y<br>Y > X | $X$ ___<br>    $Y$ ___ | X Precede Y |
| X m Y<br>Y mi X | $X$ ___<br>  $Y$ ___ | X meets Y |
| X o Y<br>Y oi X | $X$ ___<br>  $Y$ ___ | Overlaps X by Y |
| X s Y<br>Y si X | $X$ ___<br>$Y$ _____ | X starts Y |
| X d Y<br>Y di X | $X$ ___<br>$Y$ _____ | X during Y |
| X f Y<br>Y fi X | $X$ ___<br>$Y$ _____ | X finishes Y |
| X = Y | $X$ ___<br>$Y$ ___ | X is equal to Y |

### B. DBN as a mixture of several probabilistic structures

Dynamic Bayesian networks generalize hidden Markov models (HMM) and linear dynamical systems (LDS) by representing the hidden states (and seen) as state variables, with complex interdependencies. The HMMs are used to represent discrete states and the LDS are used to represent states (variables) continuous. The combination of these two structures to create a mixed-state DBN. This type of model was introduced and applied to the recognition of human gestures **[9]**

### C. Pure Probabilistic DBN

In this section we consider a DBN as a graph whose nodes represent states and arcs represent conditional dependencies (causal) between states of a band as well as temporal dependencies between the states belonging to two consecutive time slices

*1) Extension of BNs toward DBNs*

A static Bayesian network can be extended in many ways to represent temporal process. These extensions can be classified into five categories:

1- Adding the history of a node to explicitly express the temporal aspect in the Bayesian network.
2- Select from a library of pre-developed Bayesian network, the RB appropriate to the current state.
3- Changing dynamics of the network structure.
4- Repeat the traditional network for each time step by introducing Bayesian networks to represent events.
5- Repeat the classical Bayesian network by adding arcs representing the time dependencies of a time slice to another.

The networks of the first category may be regarded as mere static BNs which is added an additional node to represent past information in time. The second class of Bayesian networks is the object of an idea that has been used in early work on DBNs by Singhal et al. They use a bunch of BNs (COBRA) developed locally and every time the system selects the Bayesian network corresponding to its beliefs about the current state of real objects studied, hence the dynamic (temporal) aspect of this class

We will describe in more detail, the three other types of extension of BNs to the DBN:

*2) Dynamic change in the structure of DBNs*

Changes in the structure of a DBN can be:
- Changing network settings (values of the table of conditional probabilities CPT) of a time slice to another
- Adding or deleting new nodes and / or arcs to the structure of BN.

The structural changes of a DBN (addition or deletion of edges or nodes) is a complex problem and can not be generalized easily. In the following, we are interested in changing parameters (CPT) system.

In **[10]** Zweig and Russell presented a model that uses decomposition techniques to represent dynamic situations real. These dynamic processes can be decomposed into several sequences. Such decomposition can be used in speech recognition or recognition of manuscripts. They found it more suitable to represent dynamic processes (temporal) creating a RB (a subnet) at each stage in the evolution of the process to model the whole process by a single BN. Each sub-network must be learned from observations at the appropriate time.

*3) DBNs for events representation*

In such networks, we use information obtained from states belonging to two consecutive time slots in order to deduce the events that took place between the two points of time. Structure of these networks is presented in the following diagram:

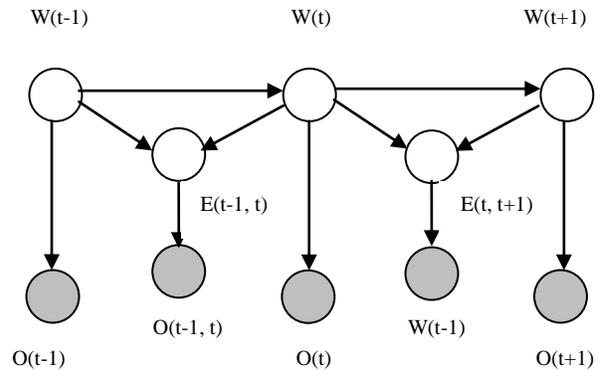

Figure 1. General structure of a dynamic belief network

In such networks, there are three types of nodes W, O and E which represent respectively:

- The random variables (corresponding to states of the real process)
- Observations
- Events

Dynamic Bayesian networks are a repetition of the traditional network in which we add a causal link (representing the time dependencies) of a time step to another. The network topology is the same for the different time slots. Arcs and probabilities that form these models have the same interpretations as for a statistical system based on a classic SNL. Thus, a DBN is completely defined by giving the couple $(B_1, B_2)$, with:

- $B_1$ is a BN which defines the a priori probability $P(X_1)$ (initial state)
- $B_2$ is the temporal Bayesian Network with two time slices (2TBN: two-slice Temporal Bayes Net) which defines $P(X_t, X_{t-1})$ using a directed acyclic graph DAG as follows:

$$P(X_t/X_{t-1}) = \prod_{i=1}^{N} P(X_t^i / C(X_t^i))$$

Where $X_t^i$ represent le i$^{th}$ node at time t and $C(X_t^i)$ is the parent of $X_t^i$ in the graph.

## V. FROM HMMs TO DBNs

The main difference between the HMM and dynamic Bayesian networks is that in an RBD the hidden states are represented as distributed by a set of random variables

...
...

$(X_t^1, X_t^2, \ldots X_t^n)$. Thus, in an HMM, the state space consists of a single random variable $X_t$.

Figure 2 shows a HMM represented in its graphical form with a dynamic Bayesian network. The gray nodes represent observed nodes and nodes in white are the hidden nodes.

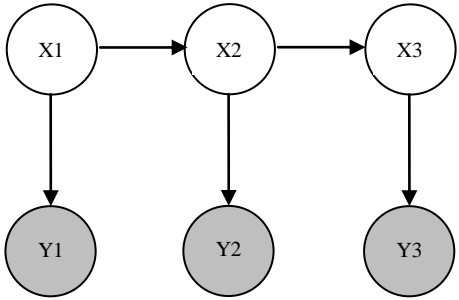

Figure 2. HMM represented as an instance of RBD unrolled over three time steps

In Figure 2, following the notations used in the literature on the HMM, the node $X_1$ represents the initial state $\pi$ with $\pi(i) = p(X_1 = i)$. The transition matrix is represented by tables of transition probabilities between nodes $X_t$ et $X_{t+1}$ with

$$A(i,j) = P(X_t = j / X_{t-1} = i)$$

Finally, the observation matrix is found in probability tables between nodes $X_t$ and $Y_t$ with

$$B(i,j) = P(Y_t = j / X_t = i)$$

Thus, the specification of an HMM as a dynamic Bayesian network is simply given by the probability tables for $P(X_1)$, $P(X_t/X_{t-1})$ et $P(Y_t/X_t)$. Assuming that the model is invariant over time (transition matrix and observation are fixed over time) then the givening of $P(X_1), P(X_2/X_1)$ et $P(Y_1/X_1)$ are sufficient.

The major advantage of dynamic Bayesian networks over HMM is that it is very easy to create alternatives to HMM simply giving another structure more or less complex DBN. The formalism and algorithms remain the same **[11]**. If you change the tables of probability distributions (discrete tables) by continuous distributions (eg Gaussian), then it also becomes possible to represent models based on Kalman filters **[12]**. It is also possible to combine these different models simply by hanging them DBN and thus provide more complex models.

## VI. REPRESENTATION OF HMMS AS DBNS

There are several variants of HMM, which were proposed in response to specific classes of problems and to overcome limitations in traditional HMMs.

In this section, we will present the variations of the most widely used HMM (shown in Figure 3). The coupled HMM (Figure 3 (a)) is probably the most natural structure, which can process, simultaneously and with good efficiency, multiple data streams from the same observations. For this, we will briefly introduce other representations and will be presented in more detail the coupled HMMs in the next section.

Figure 4 (b) is a specific coupling of HMM described in [13] as an event coupled HMM. The motivation for this representation is to model a class of loosely coupled time series where only the occurrence of events are coupled in time. The representation of events coupled with HMMs is obviously limited by its narrow structure and this structure is for a very specific class of applications.

Input / Output HMM (Figure 4 (c)) [15] represents a promising alternative to the use of a hidden Markov model. This variant allows to map an input sequence and output sequence. The main difference with traditional HMMs is indeed the first is the distribution $P(y_1^T)$ of the output sequence $y_1^T = \{y_1, y_2 \ldots y_T\}$ when the second shows the distribution conditional $P(y_1^T / u_1^T)$ of the output sequence given an input sequence $u_1^T = \{u_1, u_2 \ldots u_T\}$. This allows for spot monitoring or recognition of sequences online. The inputs and outputs can be discrete or continuous, scalar or vector.

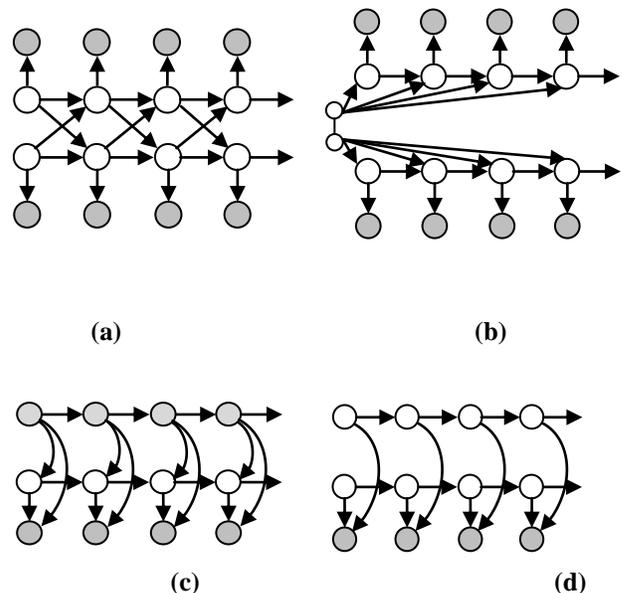

(a)  (b)

(c)  (d)

Figure 3: different variants of HMM: The empty circles represent the hidden states and the gray circles represent the observations (slightly gray circles in Figure (d) represent the input nodes). (a) coupled HMMs, (b) event coupled HMM, (c) factorial HMM, (d) input-output HMM.

The factorial HMM (Figure 3 (d)) [14] is a model used to represent systems in which the hidden states are made from a set of decoupled dynamical systems and with only one observation available.

## VII. COUPLED HMM

### A. Definition

In a coupled HMM, each hidden variable (state) is connected to his own observation. It is also connected to its two nearest neighbors in the time slice with the exception of the following variables belonging to chains border, each with a single nearest neighbor (see Figure 4).

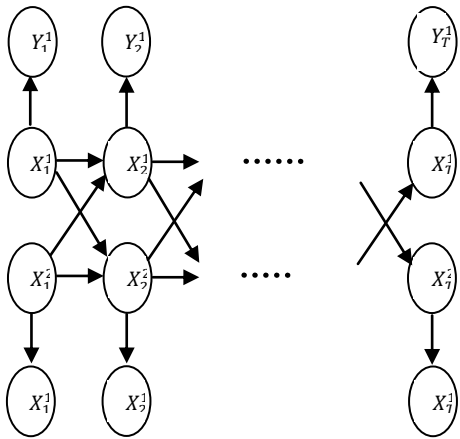

Figure 4. Coupling of two HMMs

### B. Parameters of coupled HMMs

Let a CHMM model formed with L coupled HMMs. This model is fully described giving the following parameters:
- initial probabilities:

$$\pi = \{\pi_j^{(l)}\}, \quad 1 \leq l \leq L, \quad 1 \leq j \leq N^{(l)}$$

$N^{(l)}$ is the number of states (hidden nodes) of the chain $l$

- Transition probabilities:

$$A = \{a_{ij}^{(l',l)}\}, \quad 1 \leq l, l' \leq L, \quad 1 \leq i \leq N^{(l')}, 1 \leq j \leq N^l$$
$$\sum_{j=1}^{N^{(l)}} a_{ij}^{(l',l)} = 1$$

- Probability of observation

$$B = \{b_j^{(l)}(k)\}, \quad 1 \leq l \leq L, \quad 1 \leq j \leq N^{(l')}, \quad 1 \leq k \leq M$$
$$\sum_{j=1}^{M} b_j^{(l)}(l) = 1$$

### C. Extension of the forward-backward algorithm for CHMM:

In the same way as for traditional HMMs, we use the Forward-backward algorithm to calculate $P(O/\lambda)$ in the case of L coupled HMMs. There, in this case each observation $o_t$ is a vector $(o_t^1, o_t^2, \ldots, o_t^L)$. Since L HMMs are coupled, the variables forward and backward should be defined jointly for all HMMs. In other words, we define the forward variable $\alpha_t$ as follows:

$$\alpha_t(j_1, j_2, \ldots, j_L) = P(o_1, o_2, \ldots, o_t, S_{t,J_1}, S_{t,J_1}, \ldots, S_{t,J_L}/\lambda)$$

And the backword variable $\beta_t$ as follows:

$$\beta_t(j_1, j_2, \ldots, j_L) = P(o_{t+1}, \ldots, o_T / S_{t,J_1}, S_{t,J_1}, \ldots, S_{t,J_L}, \lambda)$$

Therefore, we can calculate inductively the two variables as follows:

$$\alpha_t(j_1, j_2, \ldots, j_L) =$$

$$\begin{cases} \prod_l \pi_{j_l}^{(l)} \cdot b_{j_l}^{(l)}(o_1^l), & t = 1 \\ \sum_{i_1, i_2, \ldots, i_L} \left( \alpha_{t-1}(i_1, i_2, \ldots, i_L) \prod_l b_{j_l}^{(l)}(o_t^l) \cdot \sum_{l'} a_{i_{l'} j_l}^{(l', l)} \right), & t > 1 \end{cases}$$

$$\beta_t(i_1, i_2, \ldots, i_L) =$$

$$\begin{cases} 1, & t = T \\ \sum_{i_1, \ldots, i_L} \left( \beta_{t+1}(j_1, j_2, \ldots, j_L) \prod_l b_{j_l}^{(l)}(o_{t+1}^l) \cdot \sum_{l'} a_{i_{l'} j_l}^{(l', l)} \right), & t < T \end{cases}$$

And the likelihood function $P(O/\lambda)$ can be calculated as follows:

$$P(O/\lambda) = \sum_{j_1, j_2, \ldots, j_L} \alpha_T(j_1, j_2, \ldots, j_L)$$

### D. EM algorithm for learning parameters of CHMM

As in the case of traditional HMMs, the two basic steps of the EM algorithm as described in [3] are:
- Estimation step:
  Given the observations O, the parameters to estimate $\lambda$ and the objective function $L(\lambda; O, S)$, we construct an auxiliary function:

$$Q(\lambda; \hat{\lambda}) = E_s[L(\lambda; O, S)/O, \hat{\lambda}]$$

that represents the expectation of the objective function of all sequences of possible states, given the observations O and the current parameters estimated

- Maximization step:

In the exact EM algorithm, the role of this step is to estimate the new parameters $\hat{\lambda}_{new}$ as follows:

$$\hat{\lambda}_{new} = \arg \max_{\lambda} Q(\lambda; \hat{\lambda})$$

### VIII. MOTIVATION OF USING CHMM

According to its definition, a coupled HMM can be viewed as a collection of HMMs, one for each data stream, where the discrete nodes at time t for each HMM are conditioned by the discrete nodes at time t -1 of all HMMs linked.

The characteristics of handwritten characters can perform a joint analysis of the image of a character according to the two preferred directions: vertical ("column") and horizontal ("lines"). So we will use the coupled HMM (CHMM) to couple two HMMs: one can handle comments on the columns and the second will be used to handle comments on the lines

### IX. REFERENCES

[1] ZWEIG G., Speech Recognition with Dynamic Bayesian networks,Bayesian networks, PhD thesis, University of California, Berkeley, 1998.
[2] N. Lawrence. Variational Infererence in Probabilistic Models. PhD thesis, University of Cambridge, U.K., 2000.
[3] W. Gilks, S. Richardson, and D. Spiegelhalter. Markov chain monte carlo in practice. Interdisciplinary Statistics. Chapman & Hall, 1996.
[4] J. Pearl. Probabilistic reasoning in intelligent systems: Networks of plausible inference. Morgan Kaufmann, second edition in 1991, 1988.
[5] K. Muphy, S.Mian: Modelling Gene Expression Data using Dynamic Bayesian Natwork, Technical Report, Computer Science Division, University of California, Berkeley, CA, 1999
[6] T.A. Stephenson, An Introduction to Bayesian Network Theory and Usage, IDIAP Research Report 00-03, 2002
[7] J. Cheng, D. A. Bell, W. Liu, Learnining Belief Networks from Data: An Information Theory Based Approach, Proccedings of the sixth ACm Internatioanal Conference on Information and knowledge Management
[8] J.F.Allen, Maintaining Knowledge about Temporal Intevals, Comm. ACM, vol.26, 1983
[9] V. Pavlovic, B. Frey, and T.S. Huang, .Time series classi_cation using mixed-state ynamic Bayesian networks,. in Proc. IEEE CVPR, 1999
[10] G.Zweig, S. Russel, Compositional Modelling With DPNs, Report N. UCB/CSD-97-970, 1997
[11] Smyth P., Heckerman D. and Jordan M. Probabilistic Independence Networks for Hidden Markov Probability Models. Technical Report MSR-TR-96-03, Microsoft Research, 1996
[12] K. Murphy, Dynamic Bayesian Networks: Representation, Inference and Leaning, PhD thesis Univesity of Califonia, Berkely, 2002.
[13] T. T. Kristjansson, B. J. Frey, and T. Huang. Event-coupled hidden Markov models. In Proc. IEEE Int. Conf. On Multimedia and Exposition, volume 1, pages 385–388, 2000
[14] E. Sanchez, Réseaux Bayésiens Dynamiques pour la Vérification du Locuteur, PhD thesis Telecom Paris, Mai 2005
[15] Y. Bengio and P. Frasconi. Input-Output HMMs for sequence processing. IEEE Trans. Neural Networks, 7(5):1231–1249, September 1996.
[16] K. Jayech , MA Mahjoub "New approach using Bayesian Network to improve content based image classification systems", *IJCSI International Journal of Computer Science Issues, Vol. 7, Issue 6, November 2010.*
[17] K. Jayech , MA Mahjoub ""Clustering and Bayesian Network to improve content based image classification systems", *International Journal of Advanced Computer Science and Applications- a Special Issue on Image Processing and Analysis, May 2011.*
[18] MA Mahjoub, K. kalti "Software Comparison dealing with Bayesian networks" Lecture Notes in Computer Science (LNCS), 2011, Volume 6677, Advances in Neural Netwodks – ISNN 2011 Pages 168-177.